\documentclass[10pt,twocolumn,letterpaper]{article}

\usepackage{iccv}
\usepackage{times}
\usepackage{epsfig}
\usepackage{graphicx}
\usepackage{amsmath}
\usepackage{amssymb}
\usepackage[accsupp]{axessibility} 

\usepackage{dsfont} 
\usepackage{dutchcal} 
\usepackage{algorithm}
\usepackage{algpseudocode}%
\usepackage{caption}
\usepackage{float}
\usepackage{placeins}
\usepackage{color, colortbl}
\usepackage{stfloats}
\usepackage{enumitem}
\usepackage{tabularx}
\usepackage{xstring}
\usepackage{multirow}
\usepackage{xspace}
\usepackage{url}
\usepackage{subcaption}
\usepackage{listings}

\usepackage{xcolor}

\definecolor{codegreen}{rgb}{0,0.6,0}
\definecolor{codegray}{rgb}{0.5,0.5,0.5}
\definecolor{codepurple}{rgb}{0.58,0,0.82}
\definecolor{backcolour}{rgb}{0.95,0.95,0.92}
\lstdefinestyle{mystyle}{
    backgroundcolor=\color{backcolour},   
    commentstyle=\color{codegreen},
    keywordstyle=\color{magenta},
    numberstyle=\tiny\color{codegray},
    stringstyle=\color{codepurple},
    basicstyle=\ttfamily\footnotesize,
    breakatwhitespace=false,         
    breaklines=true,                 
    captionpos=b,                    
    keepspaces=true,                 
    numbers=left,                    
    numbersep=5pt,                  
    showspaces=false,                
    showstringspaces=false,
    showtabs=false,                  
    tabsize=2
}

\newenvironment{packed_enum}{
\begin{enumerate}
  \setlength{\itemsep}{1pt}
  \setlength{\parskip}{2pt}
  \setlength{\parsep}{0pt}
}{\end{enumerate}}

\newcommand{\topk}[1]{\underset{#1}{\operatorname{top}\operatorname{k}}\;}

\newcommand{\revision}[1]{\textcolor{black}{#1}}

\usepackage[pagebackref=true,breaklinks=true,letterpaper=true,colorlinks,bookmarks=false]{hyperref}

\iccvfinalcopy 

\def\iccvPaperID{12369} 

\ificcvfinal\fi

\begin{document}

\title{Domain Generalization Guided by Gradient Signal to Noise Ratio of Parameters}

\author{Mateusz Michalkiewicz\\
University of Queensland\\
\and
Masoud Faraki\\
NEC Labs America\\
\and 
Xiang Yu\\
Amazon \thanks{Work done while Xiang was at NEC Labs America}
\and 
Manmohan Chandraker\\
NEC Labs America,\\
University of California, San Diego \\
\and 
Mahsa Baktashmotlagh\\
University of Queensland\\
}


\maketitle
\ificcvfinal\thispagestyle{empty}\fi

\begin{abstract}
Overfitting to the source domain is a common issue in gradient-based training of deep neural networks. To compensate for the over-parameterized models, numerous regularization techniques have been introduced such as those based on dropout. While these methods achieve significant improvements on classical benchmarks such as ImageNet, their performance diminishes with the introduction of domain shift in the test set \ie when the unseen data comes from a significantly different distribution. In this paper, we move away from the classical approach of Bernoulli sampled dropout mask construction and propose to base the selection on  gradient-signal-to-noise ratio (GSNR) of network's parameters. Specifically, at each training step, parameters with high GSNR will be discarded.
Furthermore, we alleviate the burden of manually searching for the optimal dropout ratio by leveraging a meta-learning approach.
We evaluate our method on standard domain generalization benchmarks and achieve competitive results on classification and face anti-spoofing problems.
\end{abstract}
\section{Introduction}

In recent years, deep neural networks achieved remarkably good results {on several classification tasks, facilitated by regularization methods that successfully reduce over-fitting of large models to the training data.}
A simple yet powerful technique is \textit{Dropout} \cite{srivastava2014dropout}, which mutes randomly chosen activations of fully connected layers at each training iteration. A better-suited variant for Convolutional Neural Networks is {\em Dropblock} \cite{ghiasi2018dropblock}, which masks contiguous regions of feature maps with spatial correlation.
\begin{figure}[h]
    \centering
     \includegraphics[width=\linewidth]{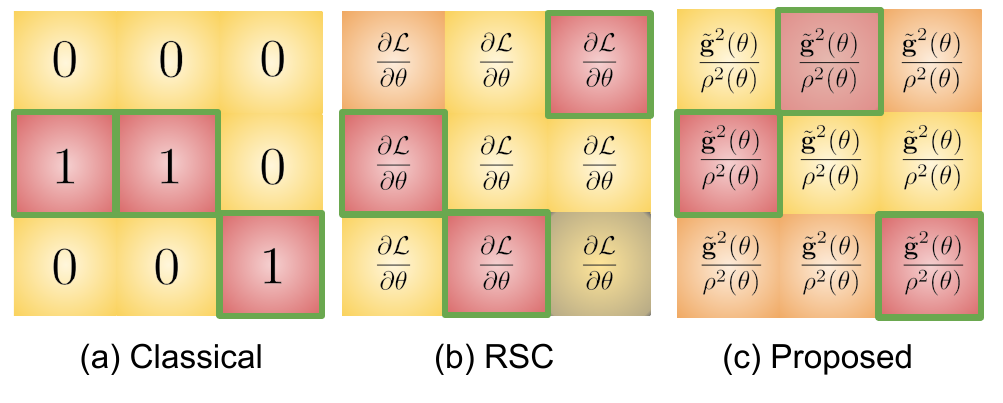}
    \caption{Different approaches to dropout mask construction. Left to right: (a) classical approach where values are sampled from a Bernoulli distribution \cite{srivastava2014dropout}; (b) RSC \cite{huang2020self}, where parameters with highest gradients are masked ; (c) ours, where parameters with highest gradient-signal-to-noise ratio are masked. The amount of discarded parameters is controlled by a dropout ratio $p$ which is typically manually chosen, whereas in our approach it is automatically learned.}
    \label{fig:gsnr_intro}
    \vspace{-4ex}
\end{figure}
While existing regularization techniques achieve great results, their success is based on the underlying assumption that the train and test data follow similar distributions. A more practical scenario, however, is presented in a {\em domain generalization} setting, where there is a distribution shift between the train and test set~\cite{simon2022generalizing,zhou2021mixstyle}. Here, models equipped with classical regularization techniques often fail to generalize their inference to unseen examples. \\

The goal of our work is to build a model that is robust to the domain shift and performs equally well on both source and unseen test domains. We build our model on two observations. First, models with high Gradient Signal to Noise Ratio (GSNR), defined as the ratio of squared mean over variance of parameters gradients on a particular data distribution, exhibit a smaller generalization gap \ie their performance does not drastically decrease when evaluated on the unseen data \cite{liu2020understanding}. Second, by iteratively dropping the most predictive parameters, the model is forced to learn less dominant features which might correspond to domain-invariant features, thus improving performance on unseen domains \cite{huang2020self}.

In light of this discussion, we carefully design a dropout strategy to drop parameters with highest gradient-signal-to-noise ratio in each training step which we illustrate in Figure \ref{fig:gsnr_intro}. As a result, the overall GSNR of the model improves which leads to a better generalization performance. Furthermore, we have observed that different blocks of neural network favour different dropout ratios. Therefore, we replace the standard approach of applying a fixed and manually chosen dropout probability $p$ by leveraging a learning-to-learn technique  \cite{finn2017model,li2017meta} to learn the dropout probability for each neural network block. Lastly, we validate our approach through extensive experiments on benchmark domain generalization datasets on classification and face recognition tasks, and show that our approach outperforms all the baselines including the ones based on {a standard} dropout strategy. More specifically, our GSNR-guided dropout is complementary to the the recent method of Representation Self-Challenging (RSC) ~\cite{huang2020self} which drops out features based on gradient magnitudes with a fixed probability. We perform extensive experiments on widely used domain generalization datasets \ie, DomainNet~\cite{peng2019moment}, OfficeHome~\cite{venkateswara2017deep} and PACS~\cite{li2017deeper} and the OCIM benchmark for face anti-spoofing consisting of 4 different datasets: OULU-NPU ~\cite{boulkenafet2017oulu}, CASIA-FASD \cite{zhang2012face}, MSU-MFSD \cite{chingovska2012effectiveness}, and REPLAY-ATTACK \cite{wen2015face}. The results show that our method consistently leads to improvements compared to the existing methods.

To summarize, our contributions are three-fold:
\begin{packed_enum}
    \item  We introduce a novel dropout strategy based on GSNR, that can be easily incorporated in {any standard convolutional} neural network architecture.
    
    \item We alleviate the problem of choosing optimal dropout ratios through {a novel} meta-learning framework. 
    
    \item We empirically validate our approach on a number of domain generalization benchmark datasets for object classification and face anti-spoofing tasks.
    
  \end{packed_enum}

\label{sec:intro}

\section{Related Work}
\label{sec:related}

\subsection{Domain Generalization} 
Domain Generalization (DG) algorithms seek to generalize beyond several source domains while reducing the distributional shift to target domains. The goal is to learn a model that is generalizable to any unseen domain. DG has been studied from a number of different angles.

A straightforward approach proposed by ~\cite{Vapnik1998}, is to utilize the empirical risk minimization strategy and train the DG model with multiple source domains. This standard idea which is borrowed from the fully supervised learning strategy is also effective for domain generalization. Another very common approach is domain alignment where the goal is to extract a domain invariant feature representation \cite{muandet2013domain, faraki2021cross}. 

Another approach is by decoupling domain agnostic features from domain specific ones as done in \cite{li2017deeper}. To boost generalization ability, Zhou \etal~\cite{zhou2021mixstyle} suggest that a style transfer strategy captured by the bottom layers of a network, be applied to the feature statistics. This is motivated by the fact that visual domains are mostly related to image styles. 

Patch shuffling input images \cite{noroozi2016unsupervised} or parsing them through random convolutions \cite{xu2020robust} are one of many examples of data augmentation approaches to DG. Domain shift may also be handled by training a separate classifier for each source domain and using the one that best fits the target domain \cite{xu2014exploiting}. 

Learning-to-learn techniques~\cite{finn2017model,li2017meta,liu2022learning}
emulate the domain shift during training by constructing meta-train and meta-test sets, assuming that good performance on meta-test set will translate well to unseen domains. An extension is suggested in \cite{zhang2020arm} to adapt the empirical risk minimization loss to a context model with a learning to learn method.

\subsection{Dropout Regularization}

A commonly used regularization technique to prevent overfitting is Dropout \cite{srivastava2014dropout}, where a binary mask is sampled from a Bernoulli distribution at each training iteration and applied to mute random parameters. Further, to compensate for the masked parameters, the outputs are scaled by $\frac{1}{1-p}$, where $p$ is the probability that a given parameter is muted. As a result, networks can learn alternative pathways to improve their predictive abilities on unseen data.

This simple procedure inspired numerous other dropout-based approaches. Popular examples are: DropPath \cite{larsson2016fractalnet} where an entire layer is muted, SpatialDropout \cite{tompson2015efficient} where dropout is applied channel-wise, CutOut \cite{devries2017improved} where random patches of input images are dropped, or AlphaDropout \cite{klambauer2017self} which is also applied to input images while preserving the original mean and standard deviation.
An approach closely related to our method is DropBlock \cite{ghiasi2018dropblock} which drops out square patches of feature maps of a given size and is better-suited than Dropout for Convolutional Neural Networks as it is more efficient in muting semantic information.

These techniques showed improvement in the generalization capabilities of neural networks as long as the train and test examples follow the same distribution. However, they are no longer effective when evaluated under a domain generalization setup where a significant domain shift is introduced \cite{zhou2021domain}. Here, zeroing the most predictive parts of features maps, such as those with highest activation values or gradients \cite{huang2020self}, has yielded better results.

\section{Methodology}\label{sec:method}

\subsection{Notations}

In the next sections we will use the following notations. Let $x$ and $y$ denote images and their corresponding labels sampled from a data distribution $\mathcal{Z}$. Let $f_\theta(\textbf{x}, M)$ denote a modified ResNet18 network \cite{he2016deep} parameterized by $\theta$, where a tensor $M$ is applied to each activation block. Let $m^{(i)}_j$ denote $j$-th element of matrix $M^{(i)}$. Finally, let $\text{sort}(\cdot)_k$ and $\topk{\text{ }}(\cdot, k). $ denote $k$ largest elements and $k$-th largest element respectively.


\subsection{Background}
Understanding why neural networks generalize is a fundamental problem. A number of works study the generalization ability of gradient-based methods \cite{rahaman2019spectral,zhang2021understanding,li2020understanding}. Recent wors introduce the concept of network stiffness, where alignment of loss gradients is linked to generalization \cite{fort2019stiffness}. More recently, two metrics are proposed to study the generalization gap~\cite{li2020understanding}: gradient-signal-to-noise ratio (GSNR) and one-step-generalization ratio (OSGR). Given a network $f$, loss function $\mathcal{L}$, images $x$, and their corresponding labels $y$ sampled from a data distribution $\mathcal{Z}$, the GSNR of a parameter $\theta$ is defined as the ratio between parameters' mean gradients and the corresponding variance, with respect to the loss function, as below:
\begin{equation}
    r(\theta) = \frac{\Tilde{\textbf{g}}^2(\theta)}{\rho^2(\theta)},
\end{equation}
where:
\begin{equation}
\begin{split}
\Tilde{\textbf{g}}(\theta) = E_{(x,y)\sim \mathcal{Z}}(\frac{\partial \mathcal{L}(f(x, \theta), y)}{\partial \theta}), \\
 \rho^2(\theta) = \text{Var}_{(x,y)\sim \mathcal{Z}}(\frac{\partial \mathcal{L}(f(x, \theta), y)}{\partial \theta}).
\end{split} 
\end{equation}

Given an empirical training loss $L[D]$ on dataset $D$ and an empirical testing loss $L[D']$ on dataset $D'$:
\begin{equation}
\begin{split}
L[D] = \frac{1}{n} \sum_{i=1}^{n} \mathcal{L}(f(x_i, \theta), y), \\
L[D'] = \frac{1}{n} \sum_{i=1}^{n} \mathcal{L}(f(x'_i, \theta), y'),
\end{split}
\end{equation}

we can define OSGR as:
\begin{equation}
    \textbf{R}(\mathcal{Z}, n) = \frac{E_{D, D'\sim \mathcal{Z}^n}(\Delta L[D'])}{E_{D\sim \mathcal{Z}^n}(\Delta L[D])},
\end{equation}
where $n$ denotes the size of datasets $D$ and $D'$, respectively.
Note that OSGR of 1 means that the performance gap between train set and test set is 0, which indicates perfect generalization.
\revision{Following \cite{liu2020understanding} we can rewrite $\textbf{R}(\mathcal{Z}, n)$ as: }
\begin{equation}
    \textbf{R}(\mathcal{Z}, n) = 1-\frac{\sum_j \rho_j^2}{n\sum_j(\frac{1}{n}\rho_j^2 + \Tilde{\textbf{g}}_j^2)},
\end{equation}
\revision{which shows OGSR's dependence on the magnitude of the gradients of network's parameters $\Tilde{\textbf{g}}$ and their corresponding standard deviations $\rho$. Assuming the learning rate is small, OGSR can be expressed as:}

\begin{equation}
    \textbf{R}(\mathcal{Z}, n) = 1-\frac{1}{n}\sum_j W_j \frac{1}{r_j + \frac{1}{n}},
    \label{eq_gsnr_osgr}
\end{equation}
\revision{where the weights $W_j=\frac{E_{D,\sim \mathcal{Z}^n}(\Delta L_j[D])}{E_{D\sim \mathcal{Z}^n}(\Delta L[D])}$ sum up to 1. 
 This shows that high GSNR of a model correlates with high OGSR, \ie, strong generalization. However, no clear way of enhancing the model's GSNR is presented.
}

\subsection{Proposed Approach}
The goal of domain generalization is to design a model that would demonstrate good generalization capabilities towards unseen domains. One way of measuring the generalization ability of a neural network is by looking at their gradient-signal-to-noise ratio (GSNR). In~\cite{liu2020understanding}, a quantitative relationship is developed between parameter's gradient-signal-to-noise ratio and the ability of the model to generalize well to unseen examples. Since the proposed ratio between mean and variance of the gradients cannot be readily optimised, we propose to enhance the model's GSNR by iteratively zeroing parameters that exhibit high GSNRs.

As noted in \cite{zhou2020domain}, simply muting random parameters during training is an ineffective strategy in a domain generalization setup. However, muting the most significant features can indeed improve cross-domain generalization. The importance of the most predictive features can be measured by looking at the magnitude of the gradients, as studied in \cite{huang2020self}. We instead focus on the parameters with high GSNR and combine it with the DropBlock technique to improve generalization capabilities and outperform \cite{huang2020self}, as well as domain generalization baselines.
In Section \ref{sec_db_gsnr}, we describe the proposed procedure in detail.

We also observe that the optimal dropout ratio varies across different ResNet blocks and different domains. A trivial solution is to find those ratios through parameter grid search. This approach, however, can be infeasible as the number of possible configurations grows exponentially with increasing depth of the network. To alleviate this computational burden, alternative dropout strategies have been proposed~\cite{huang2020self} which are only applied to a single ResNet block. To improve on the aforementioned sub-optimal strategies, we propose to learn parameters which modulate the amount of activations to be muted through a meta-learner. We describe this approach in Section \ref{sec_meta_gsnr}.

Finally,  a comparative analysis in Section \ref{sec:comparersc} highlights the differences between our approach and the one of~\cite{huang2020self}.

\subsubsection{GSNR-Guided DropBlock}
\label{sec_db_gsnr}

Our approach follows a dropout regularization procedure where, in each training step, we construct a dropout mask that mutes a subset of activations. Concretely, the forward pass consists of the following three steps:
\begin{enumerate}
    \item First, we calculate the gradients of the loss function with respect to the parameters of $i$-th ResNet block.
    \begin{equation}
       \textbf{g}_i^{(1)} = \frac{\partial \mathcal{L}(f_\theta(\textbf{x}, \mathds{1}),\textbf{y}))}{\partial\theta_{\text{block}_i}}.
    \end{equation}
    At this stage, an identity tensor $\mathds{1}$ is used to compute the logits.
    Then, we calculate the gradient-signal-to-noise ratio for each parameter $\theta_j$:
    \begin{equation}
    \label{eq_db_gsnr}
       r_j = \frac{\text{E}_{(x,y)\sim \mathcal{Z}}(\textbf{g}(x,y,\theta_j))}{\text{Var}_{(x,y)\sim \mathcal{Z}}(\textbf{g}(x,y,\theta_j))}.
    \end{equation}
    Here, the mean and variance of the data distribution $\mathcal{Z}$ is approximated by the mean and variance within the current batch. Then, we construct a binary mask where parameters with largest GSNR are zeroed: 
    \begin{equation}
    \label{eq.m1}
        m_j^{(1)} = \begin{cases} 
        1,& \text{if } r_j \geq \tau \\
        0,              & \text{otherwise}
    \end{cases},
    \end{equation}
    with the threshold $\tau$ being the $k$-th largest GSNR value in ResNet block $i$:
    \begin{equation}\label{eq.tau}
      \tau_i = \topk{j}(r_j, k).          
    \end{equation}

    Following common practice, we design a mask identifying whether muting the parameter should occur:
    \begin{equation}
    \label{eq.m2}
       m_j^{(2)} \sim \text{Bernoulli}(p_\text{gsnr}).
    \end{equation}
    \item We combine $M_j^{(1)}$ and $M_j^{(2)}$ to select which activations should be set to 0:
    \begin{equation}
       M = M^{(1)} \times M^{(2)}.
    \end{equation}
    \item Finally, we compute the gradients of the loss function with respect to all parameters:
    \begin{equation}
       \textbf{g} = \frac{\partial \mathcal{L}(f_\theta(\textbf{x}, M),\textbf{y}))}{\partial\theta}, 
    \end{equation}
    and feed them to the optimizer.
\end{enumerate}

Our approach is summarized in Algorithm \ref{db_gsnr_alg}. As described earlier, this procedure requires manually selecting the dropout ratio, (equation \ref{eq.tau}) which can be prohibitively expensive as different ResNet blocks and domains favour different dropout ratios. To alleviate this problem, in the next section, we propose to use a meta-learning approach for learning the dropout ratios. 

\begin{algorithm} [t]
  \caption{DropBlock with GSNR}\label{db_gsnr_alg}
  \begin{algorithmic}[1]
  \State {$\textbf{Input}: \text{activations $A$, } \textit{block\_size},$
  \par $\gamma, \textit{mode}, \text{gradients }g=\nabla_A(\mathcal{L}), k, p $}
\If{$\textit{mode}==\textit{Inference}$}
    \State{$\text{return } A$}
\EndIf
 \State{$\text{Compute mask }M$}
 \Statex{$\text{a. }\text{Compute GSNR with eq. \ref{eq_db_gsnr}}$}
 \Statex{$\text{b. } \text{Compute threshold $\tau$ with eq. \ref{eq.tau}}$} 
 \Statex{$\text{c. } \text{Compute mask $M^{(1)}$ using eq. \ref{eq.m1}}$}
 \Statex{$\text{d. } \text{Randomly sample $M^{(2)}$ using eq. \ref{eq.m2}  }$}
 \Statex{$\text{e. } \text{Combine both masks: } M = M^{(1)} \times M^{(2)}$}
 \State {$\text{For each zero position } M_{i,j} \text{, create a spatial square}$}
 \Statex{$\text{mask with the center being } M_{i,j} \text{, the width, height}$}
 \Statex{$\text{being } \textit{block\_size}\text{ and set all the values of $M$ to be zero }$}
 \State{$\text{Apply: }A = A \times M$}
 \State{$\text{Normalize the features: }$}
 \par{$A = A \times \textbf{count} (M) / \textbf{count\_ones} (M)$}
  \end{algorithmic}
\end{algorithm}

\subsubsection{ Meta-learning the Dropout Ratios}
\label{sec_meta_gsnr}

In this section, we aim to bypass the burden of the exhaustive grid search by employing learning-to-learn technique. However, our procedure described in Algorithm \ref{db_gsnr_alg} cannot be readily incorporated into a meta-learning framework: the sorting operation is not differentiable and computing $\frac{\partial\mathcal{L}}{\partial k}$ which is needed for the backward pass, cannot be handled through auto-differentiation engines (e.g., in Pytorch).

To alleviate this problem, we propose to sample the $M^{(2)}$ mask from a uniform distribution and apply a scaled hard sigmoid function $\phi$:
\begin{align}
    m_j^{(2)} &\sim \mathcal{U}[-1,1], \\
    M^{(2)} &= \phi(M^{(2)}+p),
\label{eq.m2_meta}
\end{align}
where $\phi$ is defined as:
\begin{equation}
    \phi(x) = \begin{cases} 
        0,& \text{if } x \leq -3 \\
        1,& \text{if } x \geq 3 \\
        \frac{x}{6} + \frac{1}{2},              & \text{otherwise}
    \end{cases}.
\end{equation} 
Note, that the parameter $p$ modulates the amount of activations to be zeroed, \ie describes the dropout ratio, and does not pose any issues with differentiation. Our modified procedure can now be used in a meta-learning framework which we base on \cite{li2017meta}. 
Concretely, during each training step, we randomly pick a subset of the current batch $B_i$ as the meta-training set $\mathcal{D}_\text{mtr}$. We adapt a meta-testing set construction from \cite{liu2022learning}, where we sample $k$ examples with the largest distance ($\mathcal{l}_2$ norm between logits) from $\mathcal{D}_\text{mtr}$: 
\begin{equation}
\label{eq.d_mte_constr}
\begin{split}
\mathcal{D}_\text{mte} &= \{ x_m \in B_i \setminus \mathcal{D}_\text{mtr} \mid  \\
  & \text{sort}(\max\limits_{x_j\in\mathcal{D}_\text{mtr}}\Vert f_\theta(x_m, \mathds{1}) - f_\theta(x_j, \mathds{1}) \Vert_2)_k  \}
\end{split}
\end{equation}

The meta-learning pass consists of two steps: a meta-train step and a meta-test step.
To adapt the learner to the classification task, we need to compute the gradients with respect to all ResNet blocks first:
\begin{equation}
    \mathcal{L}_{\mathcal{D}_\text{mtr}} = \frac{1}{|\mathcal{D}_\text{mtr}|}\sum\limits_{(x,y)\in \mathcal{D}_\text{mtr}} \mathcal{L}(f_\theta(\textbf{x}, \mathds{1}), \textbf{y}),
\end{equation}
    \begin{equation}
       \textbf{g}_i^{(1)} = \frac{\partial \mathcal{L}_\text{mtr}(f_\theta(\textbf{x}, \mathds{1}),\textbf{y}))}{\partial\theta_{\text{block}_i}}.
    \end{equation}
    
\begin{algorithm}[b]
  \caption{Meta GSNR}\label{alg_meta_gsnr_future}
  \begin{algorithmic}[1]
  \State {$\textbf{Input}:  \beta, \gamma, k, \text{images}, \text{labels}, f_\theta(x,M) $}
  \State{$\text{Initialize } \alpha, \mathbcal{p}$}
  \For {$\text{each epoch}$}
    \For{$\text{each batch }B_i$}
    \State{$ \mathcal{D}_{mtr} \leftarrow \text{random subset of }B_i $}
   \State{$\textbf{Meta-train:}$}
   \State{$\mathcal{L}_{\mathcal{D}_\text{mtr}} = \frac{1}{|\mathcal{D}_\text{mtr}|}\sum\limits_{(x,y)\in \mathcal{D}_\text{mtr}} \mathcal{L}(f_\theta(\textbf{x}, \mathds{1}), \textbf{y}) $}
   \For{$\text{each block $l$}$}
   \State{$       \textbf{g}_l^{(1)} = \frac{\partial \mathcal{L}_\text{mtr}(f_\theta(\textbf{x}, \mathds{1}),\textbf{y}))}{\partial\theta_{\text{block}_l}}$}
    \State{$ 
        r_{l_j} = \frac{\underset{x,y \in \mathcal{D}_\text{mtr}}{\text{mean}}g_l(x,y,\theta_{l_j})}{\underset{x,y \in \mathcal{D}_\text{mtr}}{\text{var}}g_l(x,y,\theta_{l_j})}$}
    \State{$\tau_l = \topk{j}(r_{l_j}, k)$}
    \State{$        m_{l_j}^{(1)} = \begin{cases} 
        1,& \text{if } r_{l_j} \geq \tau_l \\
        0,              & \text{otherwise}
    \end{cases}$}   
   \State{$m_{l_j}^{(2)} \sim \mathcal{U}[-1,1]$} 
   \State{$M_l^{(2)} = \phi(M^{(2)}+p)$}
   \State{$M_l =M_l^{(1)} \times M_l^{(2)}$}
   \EndFor
   
   \State{$\mathcal{L}_{\mathcal{D}_\text{mtr}} = \frac{1}{|\mathcal{D}_\text{mtr}|}\sum\limits_{(x,y)\in \mathcal{D}_\text{mtr}} \mathcal{L}(f_\theta(\textbf{x},M), \textbf{y}) $}
     \State{$\textbf{g} = \nabla_\theta \mathcal{L}_{\mathcal{D}_\text{mtr}}$}
  \State{$\theta'_i  = \theta - \alpha \circ \textbf{g}$} 
  \State{$\textbf{Meta-test:}$}
  \State{$\text{Construct $\mathcal{D}_\text{mte}$ using eq. \ref{eq.d_mte_constr} }$}
 
   \State{$\mathcal{L}_{\mathcal{D}_\text{mte}} = \frac{1}{|\mathcal{D}_\text{mte}|}\sum\limits_{(x,y)\in \mathcal{D}_\text{mte}} \mathcal{L}(f_{\theta'}(\textbf{x}, M), \textbf{y}) $}
   \State{$\mathcal{L_i} =\gamma \mathcal{L}_{\mathcal{D}_\text{mtr}}+(1-\gamma)\mathcal{L}_{\mathcal{D}_\text{mte}} $}
  \EndFor
  \State{$\textbf{Meta-optimization:}$}
  
  \State{$(\theta, \alpha, \mathbcal{p}) \leftarrow (\theta, \alpha, \mathbcal{p}) - \beta \nabla_{\theta, \alpha, \mathbcal{p}}\sum_i\mathcal{L}_i$}

  \EndFor
  \end{algorithmic}
\end{algorithm}

We then construct the dropout mask using previously described equations \ref{eq.m1} and \ref{eq.m2_meta} and use it to compute the meta-training loss and update the learners weights:


\begin{equation*}
    \mathcal{L}_{\mathcal{D}_\text{mtr}} = \frac{1}{|\mathcal{D}_\text{mtr}|}\sum\limits_{(x,y)\in \mathcal{D}_\text{mtr}} \mathcal{L}(f_\theta(\textbf{x}, M), \textbf{y}),
\end{equation*}

\begin{equation}
    \theta' = \theta - \alpha \circ \textbf{g}^{(1)}. \nonumber
\end{equation}

We calculate the meta-test loss using the updated learners weight $\theta'$:
\begin{equation}
\mathcal{L}_{\mathcal{D}_\text{mte}} = \frac{1}{|\mathcal{D}_\text{mte}|}\sum\limits_{(x,y)\in \mathcal{D}_\text{mte}} \mathcal{L}(f_{\theta'}(\textbf{x}, \textbf{g}), \textbf{y}).
\end{equation}

Finally, the meta-train and meta-test losses are combined with a $\gamma$-weighted average:
\begin{equation}
    \mathcal{L}_i = \gamma \mathcal{L}_{\mathcal{D}_\text{mtr}}+(1-\gamma)\mathcal{L}_{\mathcal{D}_\text{mte}},
\end{equation}
with the parameters $\theta$, learning rates $\alpha$ and dropout ratios $p$ being updated through the adaptation steps:
\begin{equation}
(\theta, \alpha, \mathbcal{p}) \leftarrow (\theta, \alpha, \mathbcal{p}) - \beta \nabla_{\theta, \alpha, \mathbcal{p}}\sum_i \mathcal{L}_i.    
\end{equation}

A summary of the procedure is in Algorithm \ref{alg_meta_gsnr_future}.

\subsubsection{Comparison with RSC}
\label{sec:comparersc}

Representation Self-Challenging (RSC)~\cite{huang2020self} is the closest work in the literature to our approach. RSC drops the features associated with higher gradients at each iteration. It introduces a number of extra parameters that needs tuning, such as: a parameter controlling whether the features are first pooled spatial-wise or channel-wise, or both, and the percentage of input images affected by RSC in each batch. Additionally, the data-specific dropout ratio $p$ is manually selected and only applied to the last ResNet block.

The fundamental difference between our approach and the one of RSC is the metric that is used to determine the importance of the features. While RSC looks at the magnitude of the gradients, our approach measures the gradient-signal-to-noise of the network's parameters. Moreover, our approach is much more portable as it simply overwrites the dropout mask construction in the DropBlock procedure. Lastly, in our approach, the dropout ratio $p$ is learned and applied to all ResNet blocks. Note that, learning dropout ratio in GSNR-guided dropblock allows our approach to achieve better performance than RSC.

\section{Experiments}
\label{sec:experiments}

\subsection{Experimental Setup}
In all of our experiments, we follow a leave-one-out evaluation protocol where we train our model on three source domains and evaluate on the fourth one. 
We use ResNet18 network \cite{he2016deep} as backbone which we pretrain on the ImageNet \cite{deng2009imagenet}. We base our learning-to-learn framework on Meta-SGD. Our dropout procedure is based on DropBlock \cite{ghiasi2018dropblock}, where we set the block size to 3. Our hyperparameters $\gamma$ and $\beta$ are set to 0.5 and 0.001, respectively. 
Following \cite{huang2020self}, we use SGD optimizer with weight decay
0.0005 and momentum 0.9 and we initialize the learning rate with 0.004 and decay it by 0.1 after 24 epochs.
Our batch size was set to 128.
In classification experiments, we have used an indentical codebase of \cite{huang2020self}
 and used the default 30 epochs for all
dropout-based methods (ours, RSC, SpatialDropout, etc).
To be consistent with the FAS competitiors, we have trained
our SSAN+ours model for 1200 epochs (as in done \cite{wang2022domain}) and our EPCR+ours model for 800 epochs (as
done in \cite{wang2023consistency}).

Unless noted otherwise, we follow a common practice of stacking a standard set of data augmentations. That is, we randomly apply the following procedures: crop and resize, horizontal flip, color jitter (\ie brightness, contrast, saturation, and hue), grayscale, and random convolution.

\subsection{PACS Classification}
PACS \cite{li2017deeper} is a popular domain generalization benchmark dataset where images are gathered in four different domains of: \textit{art painting}, \textit{cartoon}, \textit{sketch}, and \textit{photo}, with a challenging domain shift. Each image belongs to one of the seven seen categories: \textit{dog},
\textit{elephant}, \text{giraffe}. \textit{guitar}, \textit{horse}, \textit{house}, and \textit{person}. In total, PACS contains close to 10,000 images with over 1500 falling into each domain.  Following the original setup, we split the images from training domains
to 90\% train - 10\% val and test on the whole held-out domain. 

Results gathered in Table \ref{tab:pacs} show that our proposed method outperforms our main competitor \cite{huang2020self} on all 4 domains and improves the average performance by almost 2\%. We further demonstrate the applicability of our approach using two other standard architectures: AlexNet \cite{krizhevsky2012imagenet} and ResNet-50~\cite{he2016deep}. The results can be found in the supplementary materials.

\begin{table}[!h]
\centering\small
\captionof{table}{Classification accuracy (\%) on the PACS dataset \cite{li2017deeper}. The bold numbers indicate the best performance averaged across all domains, second best is underlined.}
\scalebox{0.89}{
\begin{tabular}{ |c|c|c|c|c|c| } 
 \hline
 PACS                                & artpaint & cartoon  & sketch  & photo    & Avg $\uparrow$  \\ \hline
 Deep All \cite{guo2021domain}       & 78.63    &  75.27   & 68.72   & 96.08    & 79.68 \\
 DMG \cite{chattopadhyay2020learning}                                 & 76.90    &  80.38   & 75.21   & 93.35    & 81.46 \\
 MMLD \cite{matsuura2020domain}                                & 81.28    & 77.16    & 72.29   & 96.09    & 81.83  \\
 L2A-OT \cite{zhou2020learning}                             &   83.30  & 78.20    & 73.60   & 96.20    & 82.80  \\
 DSON \cite{seo2020learning}                                & 84.67   & 77.65    & 82.23   &  95.87   & 85.11  \\
 MixStyle \cite{zhou2020domain}                           &  84.10   & 78.80    & 75.90   & 96.10    & 83.70  \\
 LDSDG \cite{wang2021learning}                               & 81.44   & 79.56    & 80.58   & 95.51    & 84.27 \\
 NAS-OoD    \cite{bai2021ood}                         & 83.74    & 79.69    &  77.27  & 96.23    & 84.23 \\
SFA-A  \cite{li2021simple}                            &  81.20     &77.80     & 73.70   & 93.90    & 81.70 \\
 SagNet \cite{nam2021reducing}                        &  83.58        &  77.66   & 76.30   & 95.47    & 83.25 \\
 DAML \cite{shu2021open}                                &83.00    & 78.10    & 74.10   & 95.60    & 82.70 \\
 StableNet \cite{zhang2021deep}                          &81.74     & 79.91    & 80.50   & 96.53    & 84.69 \\ 
 W2D~\cite{huang2022two} & - & - & - & - & 83.4 \\
 BatchFormer~\cite{hou2022batchformer} &84.8 & 75.3 & 81.1 &  93.6 & 83.7 \\ 
 ITL-Net~\cite{gao2022loss} & 83.9 & 78.9 & 94.8 &80.1 & 84.4\\
 Style Neophile~\cite{kang2022style} &84.41 &79.25 &94.93 & 83.27&\underline{85.47} \\
 \hline
 RSC   \cite{huang2020self}                        &   80.73 & 79.22 & 81.48 & 94.16  & 83.90    \\ 
 
 Ours                                & 83.64    & 80.03    & 84.37     & 95.32   & \textbf{85.84}  \\
 \hline
\end{tabular}
}
\label{tab:pacs}
\end{table}

\subsection{Office-Home Classification}

Office-Home ~\cite{venkateswara2017deep} is another popular benchmark dataset containing four domains: \textit{art}, \textit{clipart}, \textit{product}, and \textit{real-world}. 
It consists of approximately 15,000 images grouped into 65 unique categories, representing objects commonly found in household and office environments. Each of those categories is built of approximately 70 images. We followed the standard practice of training on all images of
training domains and testing on the held out one.
Similar to PACS evaluation setup, our model improves the average performance of RSC by over 2\% on all domains which we present in Table \ref{tab:oh}.

\begin{table}[!ht]
\centering
\caption{Classification accuracy (\%) on the Office-Home dataset. The bold numbers indicate the best performance averaged across all domains, second best is underlined. }
\scalebox{0.84}{
\begin{tabular}{ |c|c|c|c|c|c| } 
 \hline
 Office-Home                        & Art      & Clipart & Product & Real  & Avg $\uparrow$    \\ \hline
  baseline \cite{carlucci2019domain}  & 52.15 &  45.86 & 70.86 & 73.15 & 60.51 \\
 Jigen   \cite{carlucci2019domain}                           & 53.04    & 47.51   & 71.47   & 72.79 & 61.20 \\
 DSON   \cite{seo2020learning}                            & 59.37    & 45.70   & 71.84   & 74.68 & 62.90 \\
 L2A-OT \cite{zhou2020learning}     & 60.60    & 50.10   & 74.80   & 77.00 & \underline{65.60} \\

 CCSA \cite{motiian2017unified}        & 59.9 & 49.9  & 74.1 & 75.7 & 64.9 \\
 MMD-AAE \cite{li2018domain}    & 56.5 & 47.3  & 72.1 & 74.8 & 62.7 \\
 CrossGrad \cite{shankar2018generalizing}   & 58.4 &  49.4 & 73.9 & 75.8 & 64.4 \\
 SagNet  \cite{nam2021reducing}                            & 60.20    & 45.38   & 70.42   & 73.38 & 62.34 \\ 
 W2D~\cite{huang2022two}  & - & - & - & - & 63.5 \\
 BatchFormer~\cite{hou2022batchformer} & 57.8&51.0 &73.4 &75.1 & 64.3 \\
 Style Neophile~\cite{kang2022style} &59.55 &55.01 &73.57 &75.52 &\textbf{65.89} \\
 \hline
 RSC \cite{huang2020self}                         &  58.42  &  47.90       & 71.63        & 74.54      & 63.12       \\ 
 Ours                               & 59.46    &  52.81  & 73.85   & 74.98 & 65.28       \\
 \hline
  
\end{tabular}}
\label{tab:oh}
\end{table}

\subsection{miniDomainNet Classification}

Compared to PACS and Office-Home, DomainNet \cite{peng2019moment} is a significantly larger dataset spanning over 6 domains: \textit{clipart}, \textit{infograph}, \textit{painting}, \textit{quickdraw}, \textit{real}, and \textit{sketch}. A total of over half a million images is grouped into 345 categories. To facilitate faster prototyping and experimentation, miniDomainNet \cite{zhou2021domain2} has been introduced which reduces DomainNet to 4 domains: \textit{clipart}, \textit{painting}, \textit{real}, \textit{sketch}, with 126 classes and nearly a quarter million images. We divided the images into training and testing sets following the original setup~\cite{zhou2021domain2}.
The complexity of miniDomainNet significantly surpasses both PACS and Office-Home, thus making experimental validation much more meaningful.

\begin{table}[!h]
\centering
\captionof{table}{Classification accuracy (\%) on the miniDomainNet dataset. The bold numbers indicate the best performance averaged across all domains, second best is underlined.}
\scalebox{0.86}{
\begin{tabular}{ |c|c|c|c|c|c| } 
 \hline
 miniDomainNet                       & Clipart  & Painting     & Real    & Sketch    & Avg $\uparrow$  \\ \hline
 DANN~\cite{ganin2016domain}                          &   65.55       &    46.27      &  58.68       &    47.88       &  54.60     \\
 DCTN~\cite{xu2018deep}                           &     62.06  &  48.79 & 58.85 & 48.25 & 54.49 \\
 MCD~\cite{saito2018maximum}                           & 62.91 & 45.77 & 57.57 & 45.88 & 53.03   \\
 MME~\cite{saito2019semi}                           &      68.09 & 47.14 & 63.33 & 43.50 & 55.52     \\
 DAEL~\cite{zhou2021domain2}                           &     69.95 & 55.13 & 66.11 & 55.72 & 61.73      \\
 CMSDA~\cite{scalbert2021multi}                          & 71.38 & 53.76 & 66.23 & 56.24 & 61.90   \\ 
 FAUST~\cite{lee2023feature} &68.1 &52.2 &68.7 &59.1 &62.0 \\
 \hline
 
 RSC \cite{huang2020self}     & 65.34 & 59.72  & 66.70 & 58.94 & \underline{62.67}      \\
 Ours            & 69.41 & 61.48 & 66.81 & 62.83          & \textbf{65.13}      \\
 \hline
\end{tabular}}
\label{tab:mDN}
\end{table}

As showed in Table \ref{tab:mDN}, our method yet again outperforms RSC and other state-of-the-art models, establishing itself as a strong competitor in domain generalization setting.  
\revision{In the supplementary material, we present more experimental results showcasing the versatility of our approach.
Additionally, we provide further insights into the assumption that iterative removal of the most predictive parameters leads to the learning less dominant features.}
\begin{table*}[!ht]
\centering

 \captionof{table}{Classification accuracy (\%) on the OCIM dataset. The bold numbers indicate the best average performance measured by two different metrics: HTER and AUC.} 
\scalebox{0.74}{
\begin{tabular}{ |c|c|c|c|c|c|c|c|c|c|c| }

 \hline
\multicolumn{1}{|c|}{\multirow{1}[4]{*}{Method}} & \multicolumn{2}{c|}{O\&C\&I to M} & \multicolumn{2}{c|}{O\&M\&I to C} & \multicolumn{2}{c|}{O\&C\&M to I} & \multicolumn{2}{c|}{I\&C\&M to O} & \multicolumn{1}{|c|}{\multirow{1}[4]{*}{Avg  HTER $\downarrow$}} & \multicolumn{1}{|c|}{\multirow{1}[4]{*}{Avg  AUC $\uparrow$}} \\
\cline{2-9}          & \multicolumn{1}{l|}{HTER(\%) $\downarrow$}  & \multicolumn{1}{c|}{AUC(\%) $\uparrow$} & \multicolumn{1}{l|}{HTER(\%)$\downarrow$} & \multicolumn{1}{c|}{AUC(\%) $\uparrow$} & \multicolumn{1}{l|}{HTER(\%)$\downarrow$} & \multicolumn{1}{c|}{AUC(\%) $\uparrow$} & \multicolumn{1}{l|}{HTER(\%)$\downarrow$} & \multicolumn{1}{c|}{AUC(\%) $\uparrow$} & & \\ \hline
 MMD-AAE \cite{li2018domain}              & 27.08    & 83.19    & 44.59    & 58.29   & 31.58    &  75.18  & 40.98    & 63.08  &  36.05  &  69.93     \\
MADDG~\cite{shao2019multi}                 & 17.69    & 88.06   & 24.50    & 84.51   & 22.19    & 84.99   & 27.98    & 80.02  &  23.09  & 84.39 \\
DR-MD-Net~\cite{wang2020cross}             & 17.02    & 90.10   & 19.68    & 87.43   & 20.87    & 86.72   & 25.02    & 81.47  &  20.64  & 86.43   \\
RFMeta~\cite{shao2020regularized}                & 13.89    & 93.98   & 20.27    & 88.16   & 17.30    & 90.48   & 16.45    & 91.16 & 16.97 & 90.94  \\
D2AM~\cite{chen2021generalizable}                  & 12.70    & 95.66   & 20.98    & 85.58   & 15.43    & 91.22   & 15.27    & 90.87  & 16.09 & 90.83    \\
SDA~\cite{wang2021self}                  & 15.40    & 91.80   & 24.50    & 84.40   & 15.60    & 90.10   & 23.10    & 84.30  &  19.65 &  87.64   \\
DRDG~\cite{liudual}                  & 12.43    & 95.81   & 19.05    & 88.79   & 15.56    & 91.79   & 15.63    & 91.75  &  15.66 & 92.03 \\
ANRL~\cite{liu2021adaptive}                  & 10.83    & 96.75   & 17.83    & 89.26   & 16.03    & 91.04   & 15.67    & 91.90 & 15.09 & 92.23     \\
SSDG~\cite{jia2020single} & 9.63 & 92.82 & 13.38 & 90.97 & 15.15 & 91.89 & 19.89 & 86.29 & 14.51 & 90.49     \\

\hline
SSAN~\cite{wang2022domain}                       & 2.50   & 97.50 & 13.59 &93.67  & 17.62 & 83.91 & 19.15 & 87.95  & 13.21 & 90.75  \\
 SSAN + Ours               & 2.50   & 95.50 & 10.94 & 95.17  & 12.25 & 93.60 & 18.42 &  88.63 & 11.02 & 93.22 \\
 EPCR~\cite{wang2023consistency} & 7.50 & 93.73 & 9.89 & 93.61 & 11.93 & 91.96 & 16.12 & 88.62  &  11.36 & 91.98  \\
 EPCR + Ours               & 7.02 & 94.52 & 9.32 & 95.16 & 10.34 & 94.39 & 15.21 & 89.05 & \textbf{10.47} & \textbf{93.28} \\
 
 \hline
 \end{tabular}}
 \label{table.ocim_avg}
\end{table*}

\subsection{OCIM Face Anti-Spoofing}

Finally, to further demonstrate the applicability of our model, we tackle a different computer vision task of face anti-spoofing. It is a binary classification problem where the goal is to distinguish between real and spoofed faces. We validate our approach on a popular OCIM benchmark consisting of 4 different datasets: OULU-NPU ~\cite{boulkenafet2017oulu}, CASIA-FASD \cite{zhang2012face}, MSU-MFSD \cite{chingovska2012effectiveness}, and REPLAY-ATTACK \cite{wen2015face}. 
Various acquisition and presentation attack devices along with multiple lightning conditions contribute to the domain shift present in this benchmark. Each dataset provides videos of approximately 50 unique subjects which are then processed into frames. We follow the OCIM protocols proposed
in~\cite{shao2019multi} for cross-dataset testing.

Here, we base our model on Shuffled Style Assembly Network (SSAN) ~\cite{wang2022domain}. Concretely, we introduce a GSNR-guided dropout procedure into the feature generator and leave the remaining of SSAN~\cite{wang2022domain} unchanged. Similarly, we augment the EPCR~\cite{wang2023consistency} model.
Results presented in Table~\ref{table.ocim_avg} shows that our modification improves the results according to both metrics and on nearly all protocols.

\subsection{Comparison with RSC}

\revision{ To facilitate a fair comparison between RSC and our approach, we combine classical regularization method such as DropOut~\cite{srivastava2014dropout}, DropBlock~\cite{ghiasi2018dropblock} and SpatialDropout~\cite{tompson2015efficient} with different methods of dropout mask construction: based on random seeds, as done in the original approach, based on magnitude of the gradients, as done in RSC, and based on gradient-signal-to-noise ratio of the network's parameters. Ablation study presented in Table~\ref{table.ablation_rsc_ours} 
shows that dropout mask construction guided by GSNR of networks parameters outperforms other approaches by a significant margin.}

\begin{table}
\centering
\captionof{table}{\revision{Ablation study: Classification accuracy (\%) on
PACS dataset with various approaches to dropout mask construction: guided by Bernoulli samples \textit{rand}, magnitude of gradients of network's parameters \textit{mag}, and their gradient-signal-to-noise ratios \textit{GSNR}}}
\scalebox{0.72}{
\begin{tabular}{ |c|c|c|c|c|c| } 
 \hline
 PACS                                & artpaint & cartoon  & sketch  & photo    & Avg $\uparrow$  \\ \hline
 Dropout w/ rand \cite{srivastava2014dropout}  & 76.93  &  75.60   & 73.66   & 93.35 &  79.88 \\
 
 Dropout w/ mag    &80.12    & 76.49   & 80.75   & 94.67   & 83.01 \\
 Dropout w/ GSNR   & 81.44   & 78.02   & 82.33   & 94.07   & \textbf{83.96} \\
 \hline
 DropBlock w/ rand \cite{ghiasi2018dropblock}          & 76.4    & 75.4    & 69.0    & 95.9    & 79.2  \\ 
 DropBlock w/ mag  & 79.34   & 79.01   & 79.20   & 95.08   & 83.15 \\
 DropBlock w/ GSNR & 81.94   & 80.07   & 80.78   & 94.95   & \textbf{84.43} \\
 \hline
 SpatialDropout w/ rand~\cite{tompson2015efficient} & 73.82   & 78.07   & 77.49   & 93.25   & 80.65 \\
 SpatialDropout w/ mag    & 79.73    & 78.37    & 79.81    & 94.97 & 83.22 \\ 
 SpatialDropout w/ GSNR   & 79.68   & 78.79     & 79.47    & 95.32 & \textbf{83.31} \\
 \hline
 RSC-rand  & 76.80   & 77.71   & 78.11   & 94.31   & 81.73 \\
 RSC-mag  &   80.73  & 79.22  & 81.48  & 94.16  & 83.90 \\
 RSC-GSNR & 81.15    & 80.16  & 81.67  & 94.55  & \textbf{84.38} \\
 
 \hline
\end{tabular} }
\label{table.ablation_rsc_ours}
\end{table}

\subsection{Model Analysis}

\noindent\textbf{Ablation Study:} We compare various dropout strategies on PACS dataset, shown in Table \ref{table.ablation}. 
 Clearly, methods that mute activations based on \textit{most important} features outperform those relying on random selection. \revision{As reported in \cite{zhou2021mixstyle}, classical regularization methods fail to discover new patterns when domain shift has been introduced to the data.}
  Applying dropout to multiple ResNet blocks further improves the results, while best performance is obtained when dropout ratio is learned. \revision{We demonstrate the impact of intermediate masks $M^{(1)}$ and $M^{(2)}$ of our approach in the supplementary material. }

\begin{table}
\centering
\captionof{table}{Ablation study:  Classification accuracy (\%) on
PACS dataset with various dropout strategies. }
\scalebox{0.74}{
\begin{tabular}{ |c|c|c|c|c|c| } 
 \hline
 PACS                                & artpaint & cartoon  & sketch  & photo    & Avg $\uparrow$  \\ \hline
 Baseline                      & 77.0    &  75.9    & 69.2    & 96.0    & 79.5 \\
 CutOut  \cite{devries2017improved}                      & 74.9    & 74.9     & 67.7    & 95.9    & 78.3  \\
 MixUp   \cite{zhangmixup}                       & 76.8    & 74.9    & 95.8   & 66.6  & 78.5 \\
 CutMix \cite{yun2019cutmix}                 & 74.6   & 71.8    & 95.6   & 65.3   & 76.8 \\
 Manifold Mixup \cite{verma2019manifold}                  & 75.6  & 70.1    & 65.4   & 93.5   & 76.2 \\
 DropBlock (DB) \cite{ghiasi2018dropblock}                 & 76.4    & 75.4     & 69.0    & 95.9    & 79.2  \\ \hline 
 DB + GSNR                 & 81.94   & 80.07 & 80.78 & 94.95   & 84.43 \\
 DB + GSNR + grid search   & 83.34   & 80.33 & 82.56 & 95.26 & 85.37 \\
 DB + GSNR + meta          & 83.64    & 80.03    & 84.37     & 95.32   & \textbf{85.84}  \\
 \hline
\end{tabular} }
\label{table.ablation}
\end{table}

\begin{figure} [ht]
    \centering
     \includegraphics[width=\linewidth, trim=15pt 0pt 40pt 0pt, clip]{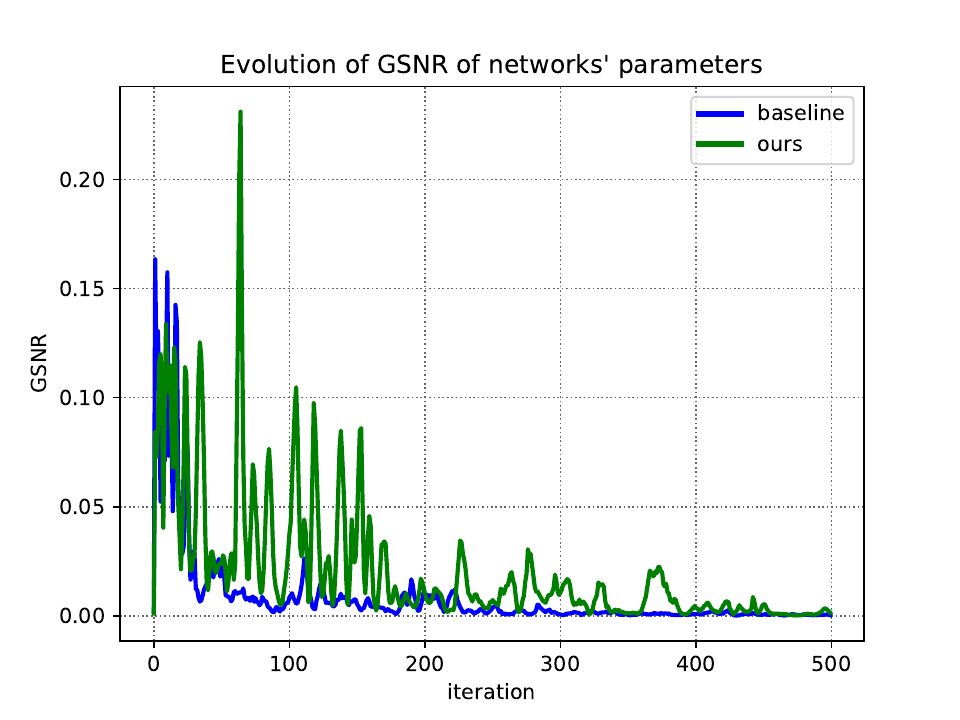}
    \caption{Our approach enhances networks gradient-signal-to-noise ratio over time, compared to the baseline.}
    \label{fig:gsnr_over_time}
\end{figure}

\noindent\textbf{GSNR of Network Parameters over Time:} To validate whether our approach enhances the overall GSNR of network's parameters over time, we register an average GSNR of all parameters over the entire training period. Figure \ref{fig:gsnr_over_time} shows that the model equipped with our procedure has higher gradient-signal-to-noise ratio compared to the baseline, confirming that our approach improves models' GSNR. \revision{GSNR evaluates the generalization ability during
training, and monitors the intermediate training process. At full convergence, the difference of signal and noise across two training models becomes negligible as the \textit{noise} is less likely to occur, reaching a similar magnitude for both models. However, the paths of how the two models are updated is quite different, resulting in different levels of generalization. Both GSNR and baseline were trained for an extended number of epochs to highlight the dynamics of overfitting.}

\begin{figure}
    \centering
     \includegraphics[width=\linewidth, trim=5pt 0pt 15pt 0pt, clip]{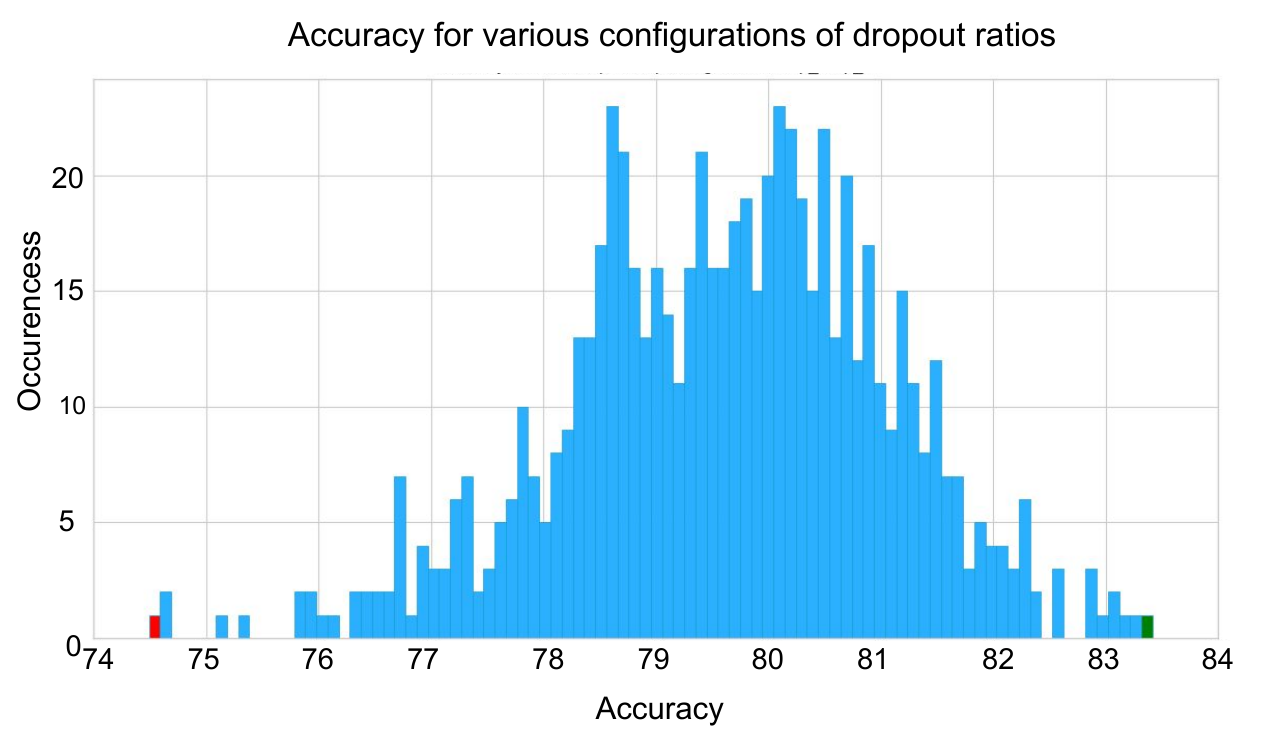}
    \caption{Accuracy distribution for different dropout ratio configurations. Worst configuration, obtained by $(0, 0.4, 0.3, 0)$ is marked in red, while best, obtained by $(0, 0, 0.2, 0.4)$ is marked in green.  }
    \label{fig:hist}
\end{figure}

\begin{figure}
    \centering
     \includegraphics[width=\linewidth, trim=15pt 0pt 40pt 0pt, clip]{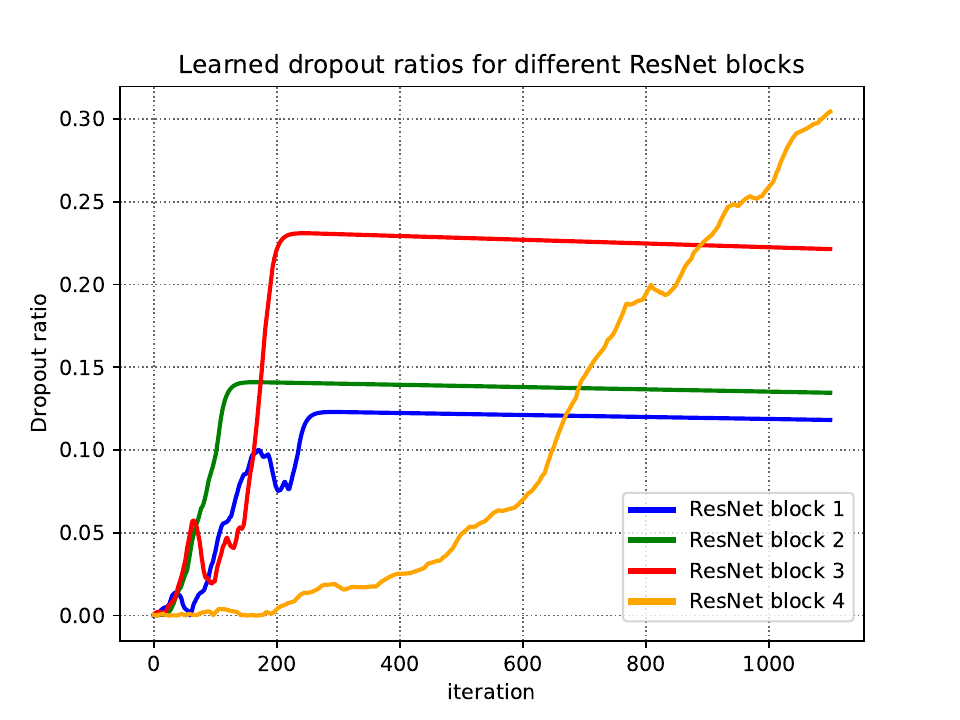}
     
    \caption{Evolution of dropout ratio over time for different ResNet blocks.}
    \vspace{-2ex}
    \label{fig:p_over_time_var_blocks}
\end{figure}

\begin{figure}
    \centering
     \includegraphics[width=\linewidth, trim=15pt 0pt 40pt 0pt, clip]{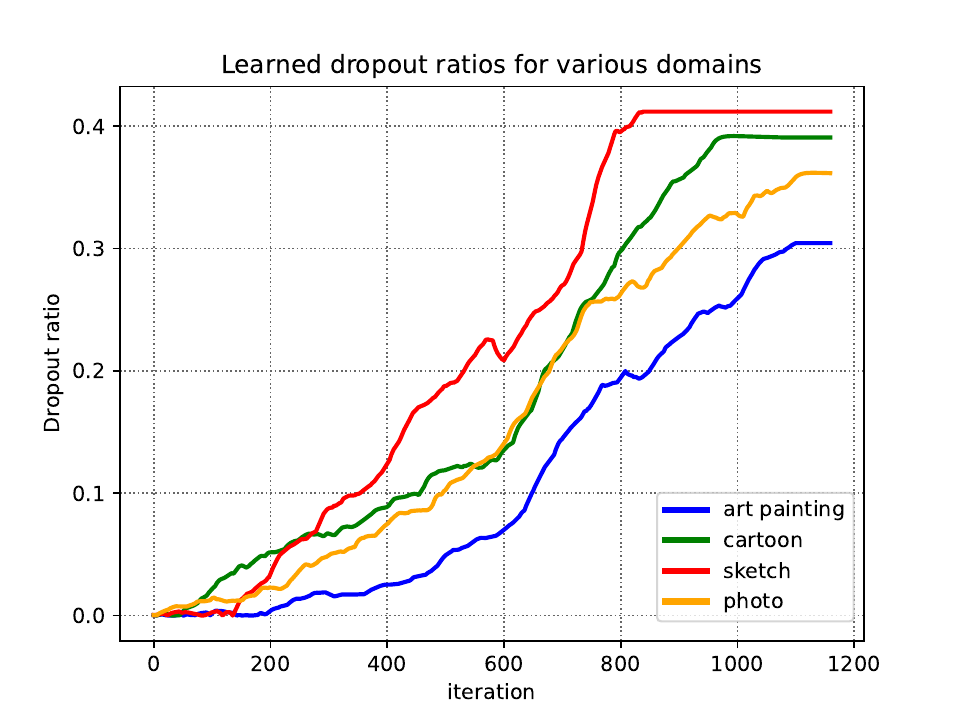}
    \caption{Evolution of dropout ratio over time for ResNet block 4 for different domains.}
    \vspace{-4ex}
    \label{fig:p_over_time_var_doms}
\end{figure}

\noindent\textbf{Learned dropout ratios:} In this section, we analyze the dropout ratios learned by our model and those found by the grid search approach. Figure \ref{fig:hist} depicts an accuracy distribution gathered by $6^4$ different configurations while Figures~\ref{fig:p_over_time_var_blocks} and \ref{fig:p_over_time_var_doms} show the evolution of dropout ratio $p$ for our meta-learning based approach. It's shown that different ResNet blocks and different domains favour different dropout ratios
supporting our claim that $p$ should be learned. 
\revision{Note that the brute force approach
is infeasible as it requires to train models with countless variations of $p$. It should rather be viewed as an oracle approach that motivates leveraging the learning-to-learn techniques for alleviating the computational burden of finding the best configuration of $p$. Finally, setting the dropout ratio to a fixed value can be far from optimal as shown in Figure~\ref{fig:hist}.} 

\noindent\textbf{Stiffness:}
We investigate the generalization capabilities of our model through the lenses of stiffness \cite{fort2019stiffness}, \ie by inspecting how much a small gradient update on one data point affects the loss on another. Specifically, the better the pairs of gradients are aligned, the higher the stiffness of the network which is linked to better generalization. We compute two variants of stiffness: intra-class and inter-class. For intra-class stiffness, where data points $x_i$ and $x_j$ share the same label $y_i = y_j$, we measure the alignment of the corresponding gradients $g_i$ and $g_j$ using the \textit{cosine} formula:
 \begin{equation}
 \label{eq.stiff_same}
     S_\text{cos}((x_i, y_i), (x_j, y_j); f) =  \mathbb{E} [\frac{\text{g}_i}{\Vert \text{g}_i \Vert } \cdot \frac{\text{g}_j}{\Vert \text{g}_j \Vert }]. 
 \end{equation}
 To compute inter-class stiffness, where labels $y_i$ and $y_j$ are different, we use the \textit{sign} formula:
 
 \begin{equation}
 \label{eq.stiff_diff}
     S_\text{sign}((x_i, y_i), (x_j, y_j); f) =  \mathbb{E} [\text{sign}(g_i, g_j)].
 \end{equation}

\begin{figure} [h]
    \centering
     \includegraphics[width=\linewidth, trim=15pt 0pt 40pt 0pt, clip]{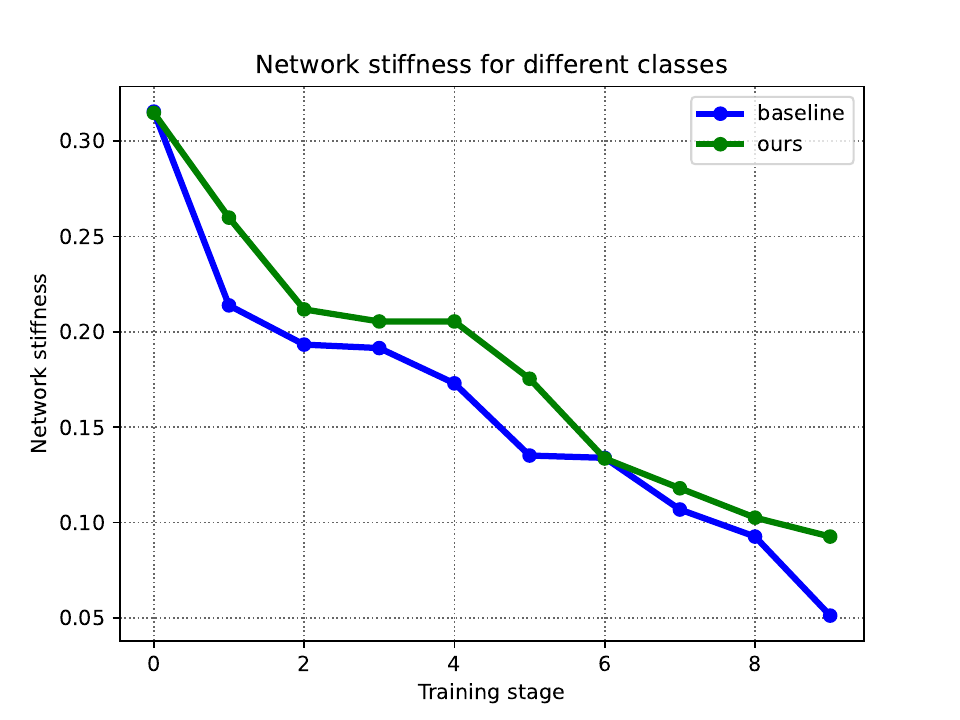}
    \caption{Evolution of \textit{inter-class} stiffness measured by Equation~\ref{eq.stiff_diff}. Here, we only consider pairs with different labels.}
    \vspace{-2ex}
    \label{fig:stiff_diff}
\end{figure}

\begin{figure} [h]
    \centering
     \includegraphics[width=\linewidth, trim=15pt 0pt 40pt 0pt, clip]{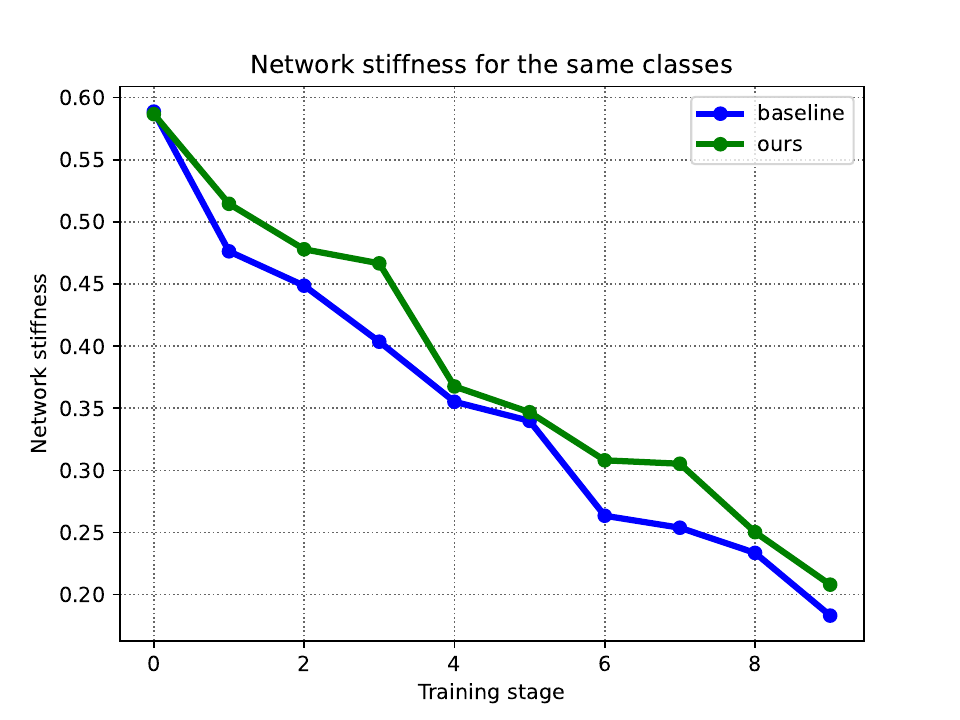}
    \caption{Evolution of \textit{intra-class} stiffness measured by Equation~\ref{eq.stiff_same}. Here, we only consider pairs with the same labels.}
    \vspace{-2ex}
    \label{fig:stiff_same}
\end{figure}

 We show that our model exhibits improved inter-class and intra-class stiffness in Figures \ref{fig:stiff_same} and \ref{fig:stiff_diff}, respectively.

\section{Conclusion and Future Work}
\label{sec:conclusion}
In this work, we introduce Meta-GSNR: a domain generalization method guided by gradient-signal-to-noise ratio of network's parameters. 
This DropBlock-based regularization procedure reduces the generalization gap by iteratively muting activations of parameters with highest GSNR values. We alleviate manual selection of dropout ratios by leveraging a learning-to-learn technique.  
Extensive experiments conducted on standard classification and face recognition benchmark datasets of PACS, Office-Home, miniDomainNet, and OCIM demonstrate the effectiveness of our approach. 

In future, we would like to investigate other gradient-based regularization metrics such as stiffness, and their effect on model generalizability.

\textbf{Acknowledgements.} This work was a part of Mateusz Michalkiewicz’s
internship at NEC Labs America. 

{\small
\bibliographystyle{ieee_fullname}
\bibliography{egbib}
}

\end{document}